\documentclass[runningheads]{llncs}
\usepackage[dvipdfmx]{graphicx}
\usepackage{color}
\usepackage{float}
\usepackage{multirow}

\begin{document}
\title{Typographic Text Generation with Off-the-Shelf Diffusion Model}
\titlerunning{Typographic Text Generation}
%

\author{KhayTze Peong\and
Seiichi Uchida\orcidID{0000-0001-8592-7566}\and
Daichi Haraguchi\orcidID{0000-0002-3109-9053}}
\authorrunning{K. Peong et al.}

\institute{Kyushu University, Fukuoka, Japan\\
\email{uchida@ait.kyushu-u.ac.jp}}

\maketitle              
\begin{abstract}
Recent diffusion-based generative models show promise in their ability to generate text images, but limitations in specifying the styles of the generated texts render them insufficient in the realm of typographic design. 
This paper proposes a typographic text generation system to add and modify text on typographic designs while specifying font styles, colors, and text effects. 
The proposed system is a novel combination of two off-the-shelf methods for diffusion models, ControlNet and Blended Latent Diffusion.
The former functions to generate text images under the guidance of edge conditions specifying stroke contours. The latter blends latent noise in Latent Diffusion Models (LDM) to add typographic text naturally onto an existing background. 
We first show that given appropriate text edges, ControlNet can generate texts in specified fonts while incorporating effects described by prompts.
We further introduce text edge manipulation as an intuitive and customizable way to produce texts with complex effects such as ``shadows'' and ``reflections''.
Finally, with the proposed system, we successfully add and modify texts on a predefined background while preserving its overall coherence. 

\keywords{Text generation \and diffusion models \and image blending.}
\end{abstract}

\section{Introduction\label{sec:intro}}

Typography is a fundamental element in textual communication. It helps convey a message, evokes specific emotions, and guides the reader through the content. Effective typographic design encompasses a thoughtful combination of carefully chosen font styles, text colors, and harmonizing backgrounds to ensure that the message is clear, legible, and easily understood.
Even subtle changes in typography may lead to significant alterations in how one perceives a design.
\par
\begin{figure}[t]
    \centering
    \includegraphics[width=0.9\linewidth]{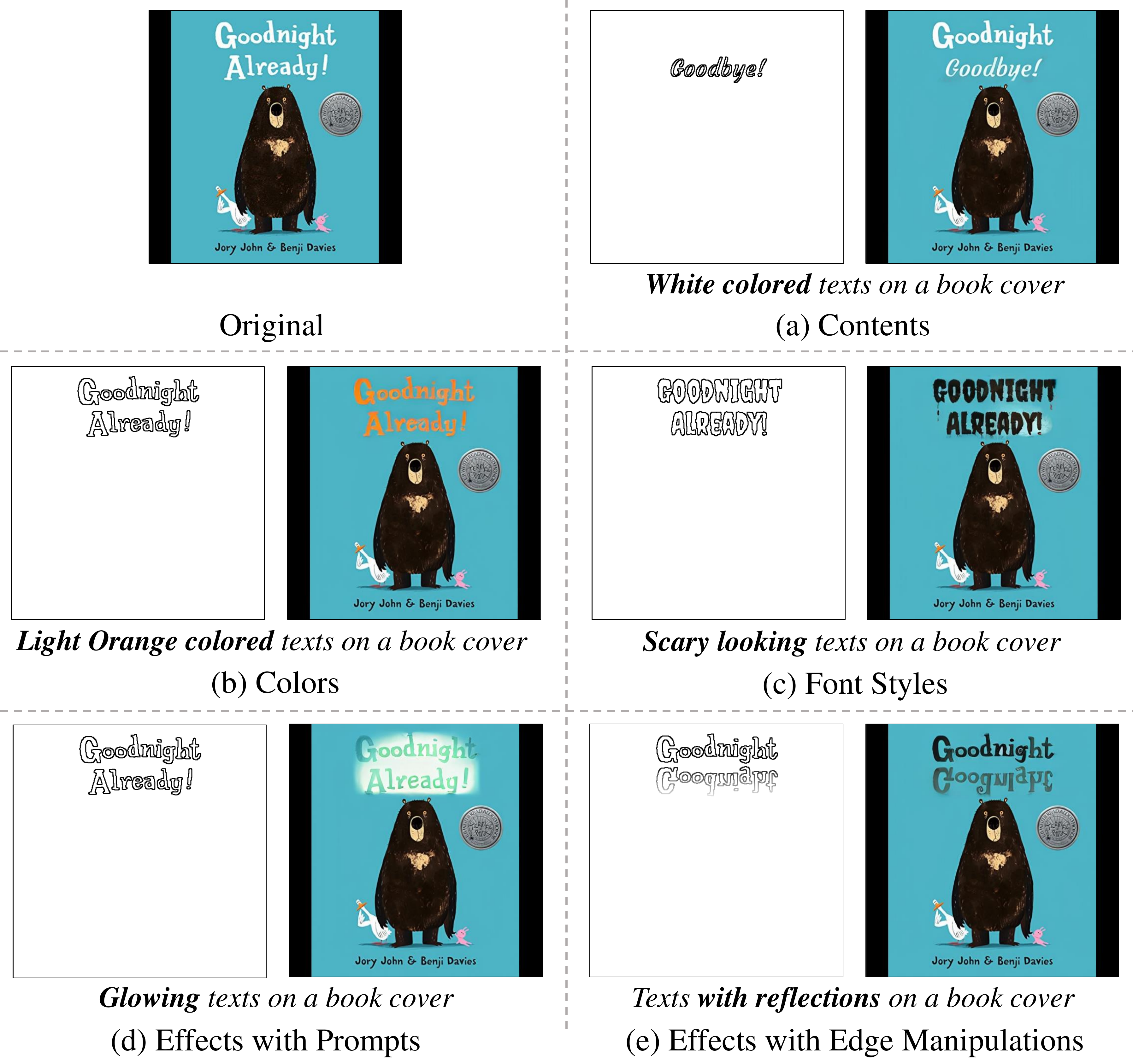}\\[-3mm]
    \caption{Examples of text generation on typographic images. The images on the left in each cell are edge conditions for generation, while the texts at the bottom are the prompts.}
    \label{fig:editing-ex}
\end{figure}

For instance, the text ``Goodnight Already'' on the book cover in Fig.~\ref{fig:editing-ex} (Original) provides insight into the content, with the chosen display font style conveying a playful impression that informs the reader about the genre of the book. 
Conversely, a change in font style in Fig.~\ref{fig:editing-ex} (c) may evoke a scary impression, even though the content remains the same.
Furthermore, adding eye-catching effects to text is also useful to emphasize certain text, as demonstrated in Figs.~\ref{fig:editing-ex} (d) and (e).
These effects may cause readers to unconsciously pay more attention to these texts than flat texts without any effect. 
\par 

Recent generative models based on diffusion models demonstrate their ability to generate not only photo-realistic images but also aesthetically pleasing text images~\cite{chen2023textdiffuser,liu-etal-2023-character,ma2023glyphdraw,yang2023glyphcontrol,zhang2023brush}.
Imagen~\cite{saharia2022photorealistic} and DeepFloyd~\cite{DeepFloyd} achieved better text image generation with prompts by utilizing Large Language Models (LLMs) as text encoders. On the other hand, Liu~\textit{et al.}~\cite{liu-etal-2023-character} focused on making text encoders character-aware to mitigate misspellings.
In addition to prompts, GlyphControl~\cite{yang2023glyphcontrol} and TextDiffuser~\cite{chen2023textdiffuser} proposed utilizing text images as additional conditions to enhance the quality of the generated texts.

However, they remain insufficient for generating typographic texts because they lack control over font style and text effects to meet the designers' demands. 
In the context of typographic designs, generative models must satisfy the following four strong requirements and one optional requirement.
\begin{enumerate}
    \item As the minimum requirement, generated texts must be legible and readable. 
    \item Positions and contents of texts can be specified by the designer.
    \item Font styles, colors, and sizes can be specified in detail to convey a particular impression intended by the designer.
    \item Texts can be added or replaced on a predefined background image because photographic and graphic elements are often predetermined. (For example, when designing the book cover of Harry Potter, we need to put the title text on a given background image showing Harry Potter.)
    Therefore, generative models need to be able to generate text while preserving the coherence of the background. 
    \item The optional requirement is a function to add various effects around texts to make texts more salient. It is often even mandatory to enhance stroke edges to better contrast individual letters from the background. 
\end{enumerate}
\par

This paper proposes a typographic text generation system that satisfies all five requirements. 
The proposed system generated the examples in Fig.~\ref{fig:editing-ex}.
Given text edges as guides and a prompt, the proposed system (a)~puts arbitrary texts at a specific position (regardless that some other texts already exist in the position or not), (b)~specifies colors, (c)~details font styles, and (d)(e)~add visual effects around texts. Through these functions, the system lowers the barriers for designers and potentially improves the efficiency of the typography design process. 
\par

Notably, our system has two ways to add visual effects: effect by prompt and effect by edge manipulations. In Fig.~\ref{fig:editing-ex} (d), the effect (glowing) is specified by the prompt. On the other hand, example (e) uses edge manipulation, enabling us to control effects more directly and intuitively. In this example, a reflection effect is realized successfully by pasting an inverted version of the original edge below it. Different from general visual objects (such as dogs and faces), texts are naturally described by edges. Therefore, edge manipulation brings about a variety of changes to the text. We utilize this nature of texts for edge-based effect manipulation. Our experiments demonstrate that diverse effects such as shadows, outlines, and 3-D appearances can also be integrated into texts with appropriate text manipulations. This function offers a fresh perspective on utilizing the edge conditions and holds promise for simplifying the process of adding effects in typographic design, particularly for non-expert designers.
\par

From a technical viewpoint, the proposed system is a novel combination of two off-the-shelf methods for diffusion models, ControlNet~\cite{zhang2023adding} and Blended Latent Diffusion~\cite{avrahami2023blended}. 
Both methods may not seem directly related to typography but, in fact, are quite useful in realizing a system that satisfies the above typographic requirements. The former functions to generate text images under the guidance of edge images specifying stroke contours, as described above. The latter blends latent noise in Latent Diffusion Models (LDM) to add typographic text naturally onto a specified background.
This combination enables adding new texts and applying additional effects to existing rasterized texts by manipulating the original text edges. We refer to the latter process as applying post-effects to texts.
These successful achievements give us hope that diffusion models are useful for typographic text generation and do not exhibit unwanted artifacts in their generated images.

Our contributions are summarized as follows.
\begin{itemize}
    \item We show that a combination of the off-the-shelf methods realizes a typographic text generation system that can satisfy multiple requirements for the practical typographic process.
    \item Through preliminary experiments in generating single-word images, we experimentally prove that controlling the proposed system with appropriate text edges and prompts generates high-quality typographic designs that display desired texts that accurately reflect the designer's intention. Specifically, we demonstrate that we can control the font style by specifying its edges and colors.
    \item We introduce edge-based text effect control in addition to prompt-based control. Specifically, manipulating text edges allows us to intuitively control various (and even complex) text effects. 
\end{itemize}

\section{Related work} 
Numerous text image generation methods have been proposed since the advent of generative models, such as generative adversarial networks (GANs). Most of them generate images of single letters~\cite{gao2019artistic,wang2020attribute2font,xie2021dg} or single words~\cite{fogel2020scrabblegan,luo2022siman} and thus do not aim to integrate the generated text with a given background image. Exceptionally, for supporting typographic designs, there were several attempts to generate text images on a background. Miyazono~\textit{et al.}~\cite{miyazono2021font} proposed text image generation that matches the specified book cover, while Gao~\textit{et al.}~\cite{gao2023textpainter} proposed text image generation for poster designs.
\par

In recent years, diffusion models have shown better performance than GANs in 
general image generation tasks~\cite{dhariwal2021diffusion}; of course, we can find many text image generation methods~\cite{chen2023textdiffuser,liu-etal-2023-character,yang2023glyphcontrol,zhang2023brush,zhu2023conditional}, which are based on diffusion models. Since text images by the original diffusion models are often illegible or unnatural, the recent models have focused on generating legible text within images. Liu~\textit{et al.}~\cite{liu-etal-2023-character} proposed using a character-aware text encoder to reduce misspellings in generated images. GlyphControl~\cite{yang2023glyphcontrol} and TextDiffuser~\cite{chen2023textdiffuser} utilize image-level conditions such as glyph images and segmentation masks to guide the generation of legible texts. 
We also can find diffusion models focusing on artistic typography~\cite{he2023wordart,IluzVinker2023,tanveer2023ds}.
\par

However, none of the existing methods satisfies even the basic requirements for generating texts in typographic works. As noted in Section~\ref{sec:intro}, the methods must generate results that accurately reflect the designer's multi-directional intents.
Table~\ref{tab:func-comparison} summarizes how state-of-the-art text generation models,  GlyphControl~\cite{yang2023glyphcontrol}, TextDiffuser~\cite{chen2023textdiffuser}, and Brush Your Text~\cite{zhang2023brush}, and ours satisfy the basic requirements for typographic works.
Neither GlyphControl nor TextDiffuser can specify font styles and text effects because they have no direct way to specify letter shapes. In contrast, Brush Your Text~\cite{zhang2023brush} can specify the shapes like ours; however, it is a model to generate whole images (including the background region of the texts) and, therefore, cannot incorporate generated texts with a pre-specified background image.
In addition, none of them proved that edge manipulation is a very promising way to add various text effects in an intuitive and accurate manner. Our system considers that edge is the key descriptor for text shape, and we fully utilize it to control the appearance of generated text images.
\par

\begin{table}[t]
\centering
\caption{Comparison of popular state-of-the-art text generation models from the viewpoint of typographic text design.}
\scalebox{.9}[.9]{
\begin{tabular}{l|ccc|c}
\hline
                                & Glyph-                                & Text-                                 &  Brush Your            & Ours            \\
                                & Control\cite{yang2023glyphcontrol}    & Diffuser\cite{chen2023textdiffuser}   &  Text\cite{zhang2023brush} &           \\
\hline
legibility of generated texts          & $\surd$             & $\surd$             & $\surd$     & $\surd$         \\
direct text content and position specification                & $\surd$             & $\surd$            & $\surd$      & $\surd$         \\
direct text style specification               & $\times$            & $\times$           & $\surd$ & $\surd$         \\
direct background image specification         & $\times$             & $\surd$             & $\times$   & $\surd$         \\
additional effect manipulation           & $\times$            & $\times$      & $\times$      & $\surd$  \\
\hline                
\end{tabular}
}
\label{tab:func-comparison}
\end{table}

With a different purpose, text image editing~\cite{DiffUTE,krishnan2023textstylebrush,Roy_2020_CVPR,wu2019editing,Yang_2020_CVPR} has been tackled recently. This is a task to modify a text (say ``90\% price down'') into a different one (such as ``80\% price down"), where the replaced character (`8') should keep the original style of `9' to be consistent with other characters. Although these methods are useful for modifying the existing typographic images by replacing a couple of characters (or words) with others, they are not useful in a more typical typographic task to put new texts on a given background image. 

\section{Preliminary Experiment: Text Image Generation with ControlNet} \label{sec:text-generation}

\subsection{How to Generate Text Images by ControlNet} 

ControlNet~\cite{zhang2023adding} is an additional neural network to incorporate spatial conditions into an existing diffusion model, especially Stable Diffusion~\cite{rombach2022high}. As spatial conditions, we can use various modalities, such as edge maps or segmentation maps. These conditions are very useful and intuitive to control the visual appearance of the generated images. For instance, with edge conditions, ControlNet can generate objects whose shapes and positions closely adhere to the provided edges. 
\par

\begin{figure}[t]
    \centering
    \includegraphics[width=0.9\linewidth]{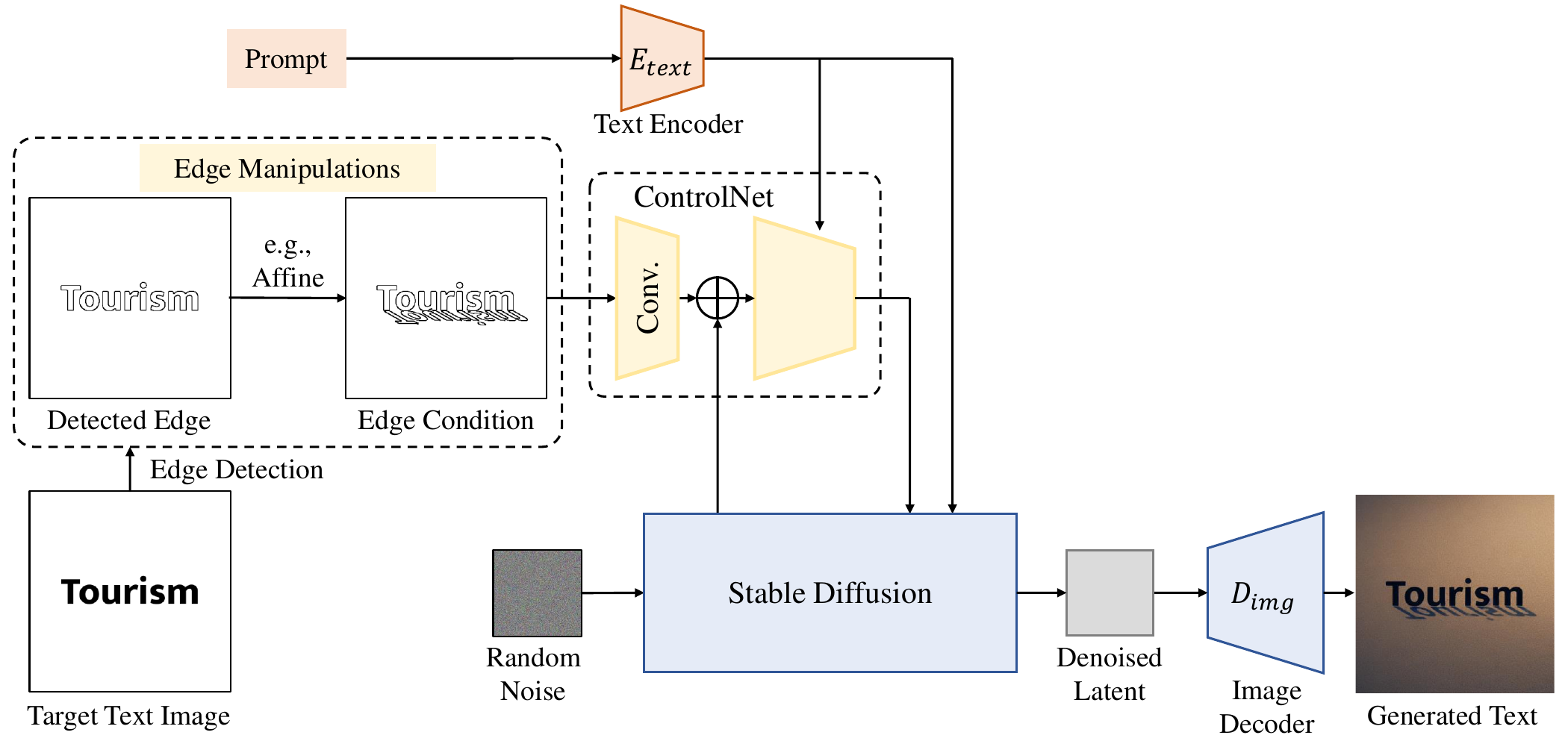}\\[-3mm]
    \caption{Overview of text generation with edge manipulation.}
    \label{fig:overview_generation}
\end{figure}

In this section, we conduct a preliminary experiment to generate single-word images with the conditions of text edges. 
Fig.~\ref{fig:overview_generation} shows an overview of the framework. A target text image is converted into an edge image (by the Canny edge detector) and then used as the edge condition for the text image generation. Then, ControlNet feeds into Stable Diffusion, and we finally get the text image that reflects the edge accurately. 
\par

Although originally proposed for generating photo-realistic images, we find that ControlNet is very appropriate for text image generation, especially for the following reasons. 
\begin{itemize}
    \item ControlNet can generate text images by accurately reflecting the edges given as a condition. This fact is advantageous because ControlNet generates text images in specific font shapes (i.e., styles). It is also advantageous to avoid strange letter shapes, which are often found in images generated by the standard diffusion models, such as Stable Diffusion~\cite{rombach2022high}.
    \item ControlNet accepts various prompts to control the appearance of resulting images. We utilize this function for adding various visual effects (such as ``3-D'' and ``shadows'') around the generated texts.
    \item We do not need to conduct any additional training for ControlNet (and Stable Diffusion) when we use them for text image generation. In fact, we conducted the following experiments with the ControlNet pre-trained with edge conditions provided in their original GitHub~\footnote{\url{https://github.com/lllyasviel/ControlNet}}.
\end{itemize}

In addition to the above reasons, we can take one more advantage of using ControlNet for typographic text generation; that is, by manipulating (i.e., modifying) edges, we can realize more various effects. The effect control by edge manipulation is direct and much more intuitive than an indirect effect control by prompt. The latter experiments in Section~\ref{sec:manipulation_generation} prove that edge manipulation actually realizes fine and complex effects while reflecting the designer's intention accurately.

\subsection{Experimental Setting} \label{sec:settings_generation}
We conducted a quantitative and qualitative experiment to confirm how ControlNet generates text images according to the user's specifications.
First, in Section~\ref{sec:quantitative_generation}, we performed OCR evaluation on the text images generated by each method, with and without font style specification, to evaluate the legibility of the generated texts in both settings.
Then, we qualitatively inspected each method's ability to reflect the specified font styles, colors, and effects in Section~\ref{sec:qualitative_generation}.

\subsubsection{Comparative Methods}
We prepared two diffusion models specialized in text generation as comparative methods:
\par
\textbf{GlyphControl~\cite{yang2023glyphcontrol}} is an extended version of ControlNet trained to generate visual texts using rendered glyph images as input condition maps. By specifying the content and text box size, they render texts on a white canvas using a fixed font as the condition for generation. This enables control over the generated texts at the layout level.

\textbf{TextDiffuser~\cite{chen2023textdiffuser}} is a two-step model specialized for text generation.
The first step uses a segmentation model to obtain a character-level segmentation mask from a text layout image (i.e., text on a white canvas). 
The second step then uses the segmentation mask obtained in step one, along with a text prompt, as input to the diffusion model to generate text images.
Here, the text layout image can either be automatically generated from prompts using a Transformer-based layout generator~\cite{gupta2021layouttransformer} or any user-specified text image. As TextDiffuser is also trained to generate texts on user-specified images, we also compared its performance with our approach on typographic text generation in Section~\ref{sec:text-editing}. \par

We do not compare our approach with Brush Your Text\cite{zhang2023brush} in the quantitative evaluation of Section~\ref{sec:quantitative_generation} because both approaches are based on ControlNet and thus expected to produce similar outputs. As clarified in Table~\ref{tab:func-comparison}, a difference between ours and \cite{zhang2023brush} in their text generation ability is edge manipulation, whose impact will be examined in Section~\ref{sec:manipulation_generation}. (Another and larger difference is the capability of merging the generated text image into a given background image; this point is also examined in Section~\ref{sec:text-editing}.)

\subsubsection{Text Generation for Evaluation}
We generated three sets of text images for each method to evaluate their typographic text generation capabilities on legibility, style specification, and effect specification. 
All images were generated based on target texts rendered to fit on white canvas images of size 512$\times$512.
The target texts used were rendered with words randomly chosen from the Synthtext~\cite{gupta2016synthetic} corpus and fonts randomly selected from a list of 3,198 Latin fonts available on Google Fonts~\footnote{\url{https://github.com/google/fonts}}.

These target text images were then used to obtain condition maps for all methods.
For our approach, we obtained text edges from the target texts using the Canny edge detector.
For GlyphControl, we rendered glyph images with the same word, size, and position as the target texts using its default font (i.e., Calibri). 
For TextDiffuser, we obtained character-level segmentation maps from the target texts using the provided segmentation model. 

\noindent
The following sets of images are generated for each method:

\textbf{Content Only} is a set of 10,000 text images generated with prompts only specifying the contents of texts to be generated. The prompt used is ``\textit{Text that reads {\tt [word]}}''. This set is used for Optical Character Recognition (OCR) evaluations to evaluate the legibility of the generated texts.

\textbf{Content w/ Font Specification} is another set of 10,000 text images generated with prompts specifying the contents and font category of texts to be generated. The categories include \textit{Serif, San Serif, Display, Handwriting} and \textit{Monospace} following the category stated in the metadata of Google Fonts. The prompt used is ``\textit{Text that reads {\tt [word]} written in {\tt [category]} font}''. 

\textbf{Content w/ Color \& Effect Specification} includes text images generated with prompts specified with various colors and effects. Types of effects used are \textit{Blue colored, Neon colored, 3-D, Glowing, Shadows} and \textit{Reflections}. For each effect, 100 text images were generated. The prompt used is ``\textit{{\tt [effect]} text that reads {\tt [word]}}'' or ``\textit{Text with {\tt [effect]} that reads {\tt [word]}}'' depending on the type of effect. This set is used only for qualitative evaluation. 

\subsection{Quantitative Evaluation of Generated Texts} \label{sec:quantitative_generation}
\subsubsection{Evaluation Metrics} \label{sec: ocr-metrics}
We conducted OCR evaluations to measure the legibility of generated texts using four metrics: precision, recall, F1-score, and normalized edit distance (NED).
All metrics are calculated for each sample and then averaged.
For evaluations, we used the open-source OCR model called PP-OCRv3~\cite{li2022pp}. 
We consider the generated text to be correct only if the recognized text perfectly matches the target text, taking into account the letter case. 
\textbf{Precision} measures the proportion of detected words that match the ground truth texts. 
\textbf{Recall} measures the proportion of ground truth texts that are actually detected in the generated images. 
\textbf{F1-score} represents the harmonic mean between precision and recall.
\textbf{Normalized edit distance (NED)} measures the edit distance between the detected text and ground truth text normalized by text length. 
Following~\cite{shi2017icdar2017}, the final NED score is calculated as one minus the average of normalized edit distance.


\subsubsection{Evaluation Results}
We quantitatively evaluated the OCR performance of both ``Content Only'' and ``Content w/ Font Specification'' for all methods and present the evaluation results in Table~\ref{table:generation-quantitative}.
From the table, ControlNet with text edge shows the best performance in almost all metrics, outperforming other models trained specifically for text generation. 
This indicates that ControlNet can generate legible texts when given appropriate text edges as conditions.
When a font category is specified, the performance of our approach remains roughly the same, while other methods experience drops in all metrics. 
Notably, GlyphControl suffers a significant drop in OCR performance. 
We observed that as both comparative methods attempt to reflect the specified font style, the legibility of generated texts decreases.

\begin{table}[t]
    \centering
    \caption{OCR evaluation of the generated texts.}
    \label{table:generation-quantitative}
    \begin{tabular}{c|cccc|cccc} \hline
                & \multicolumn{4}{c}{Content Only} & \multicolumn{4}{|c}{w/ Font Specification} \\
         Method & Precision$\uparrow$ & Recall$\uparrow$ & F1-score$\uparrow$ & NED$\uparrow$ & Precision$\uparrow$ & Recall$\uparrow$ & F1-score$\uparrow$ & NED$\uparrow$ \\ \hline
         GlyphControl                   & 0.4541          & 0.4595 & 0.4559  & 0.8114 & 0.3756 & 0.3834 & 0.3781 & 0.7633\\
         TextDiffuser                   & \textbf{0.7818} & 0.7832 & 0.7822  & 0.9192 & 0.7628 & 0.7639 & 0.7631 & 0.9141\\
         ControlNet+   & \multirow{2}{*}{0.7804} & \multirow{2}{*}{\textbf{0.8290}}  & \multirow{2}{*}{\textbf{0.7942}}    & \multirow{2}{*}{\textbf{0.9243}}  & \multirow{2}{*}{\textbf{0.8025}} & \multirow{2}{*}{\textbf{0.8225}} & \multirow{2}{*}{\textbf{0.8081}} & \multirow{2}{*}{\textbf{0.9233}}\\ 
         Text Edge      & & & & & & & &  \\\hline
         
    \end{tabular}
\end{table}
\subsection{Qualitative Evaluation of Generated Texts} \label{sec:qualitative_generation}

\begin{figure}[t]
    \centering
    \includegraphics[width=0.9\linewidth]{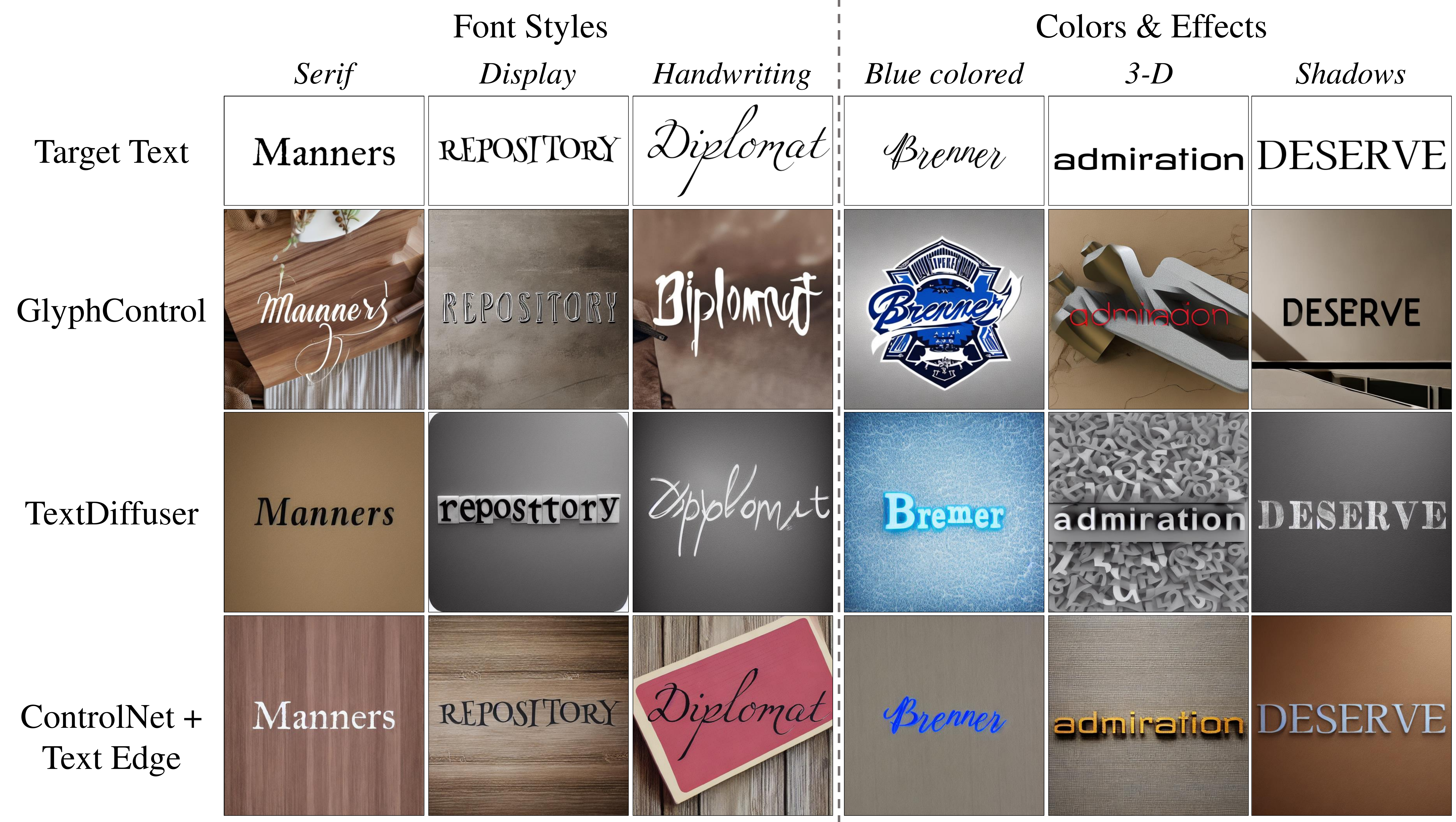}\\[-3mm]
    \caption{Examples of text images generated with different prompts. The left half displays generation results when the font style is specified, while the right half shows results when colors and effects are specified.}
    \label{fig:font-efftect-prompts}
\end{figure}

We compare all methods qualitatively on the generation results of ``Content w/ Font Specification'' and ``Content w/ Color \& Effect Specification''. 
Fig.~\ref{fig:font-efftect-prompts} displays some text images generated from target text images shown in the top row. 
The left half uses prompts specifying font, while the right half uses prompts specifying colors and effects.

\subsubsection{Font Style Specification}
Our approach of using text edges as conditions for ControlNet successfully generates texts that faithfully follow the font styles of the target texts, even with complex fonts (e.g., cursive fonts). 
In contrast, while TextDiffuser can occasionally reflect generic font styles such as Serif and Sans Serif, both comparative methods failed when the specified font styles were more complex. 
Note that even with generic font styles, TextDiffuser is unable to precisely replicate the fine strokes of the specified font, as it only uses a character-level segmentation map of the target text.

\subsubsection{Effects by Prompts}
Results on the right half of Fig.~\ref{fig:font-efftect-prompts} show that we can modify color and incorporate effects into texts by specifying prompts while maintaining the specified font style.
We observed that GlyphControl failed to reflect the specified effects other than color-related ones (i.e., \textit{Blue colored, Neon colored}). This might be due to the fact that GlyphControl was trained using datasets that predominantly contain texts without the effects specified in this experiment. 

However, prompts alone are unable to produce satisfying results for complex effects. 
For example, ControlNet only managed a weak \textit{shadow} effect, as shown in the bottom right of Fig.~\ref{fig:font-efftect-prompts}. 
While all methods failed to incorporate the \textit{reflection} effect. 
Additionally, there is no way to control the directions and strength of effects by only prompts.

\subsection{Edge Manipulation for Adding Effects around Texts} \label{sec:manipulation_generation}
In this section, we propose an intuitive method to incorporate complex effects through edge manipulations.  This approach not only adds such effects but also offers high customizability.
In Section~\ref{sec:text-generation}, we demonstrated that  ControlNet could add colors and text effects by specification of prompts. However, indirect effect control by prompt specifications is 
sufficient for neither finer nor complicated effect controls. In contrast, effect control by edge manipulation overcomes the limitation of the prompt-based control.
\par

In Fig.~\ref{fig:overview_generation}, the process of generating complex effects by edge manipulation is described. We assume a text edge image prepared from a target text image using the Canny edge detector. Then, various manipulations (e.g., affine transform, flip, etc.) are applied to the edge image. Finally, the manipulated edge is passed to ControlNet together with a prompt to produce text with the desired effect.
\par

We introduce four main edge manipulation approaches as follows:
\par
\begin{itemize}
\item \textbf{Shift} is utilized to introduce various effects such as 3-D, shadows, and repeating outlines.
We shift the edges by the desired number of pixels in the x- and y-axes. 
We then remove any edges that overlap with the original text regions. 
Finally, the original edges are reintroduced. 
\par
\item \textbf{Dilation} is for adding an outline effect around texts.
For this, we create a dilated version of the original text to obtain the outer edges and paste it around the original edges. 
The width of the outline effect is customizable by specifying the amount of dilation.
\par
\item \textbf{Flip and fade} are introduced to produce a reflection effect. 
We provide a guide for generating reflections of texts by flipping the original edge vertically and pasting it below the original edge. 
We also fade the flipped edges so that edges further from the original edges are weaker.
Through experiments, we observed that a more aesthetically pleasing reflection effect is produced when the flipped edges are faded.
\par
\item \textbf{Affine transformation} can be applied to produce a shadow effect different from Shift. 
We first rotate the original edges to the desired angle using affine transformation.
Then, similar to Shift, we remove edges that overlap with the original text regions and reintroduce the original edges.
We noticed that a stronger edge produces finer shadows, while a softer edge yields blurry shadows.
Therefore, we apply a strength multiplier of range $(0, 1]$ to the rotated edges to allow control over the sharpness of the shadows. 
\end{itemize}
\par
We emphasize that all manipulations are achieved through straightforward image-processing approaches, and various effects can be incorporated by simply adding a prompt that matches the manipulated edges.
This simplicity and customizability facilitate the creation of typography text for non-expert designers.
Additionally, we believe there is further potential for edge manipulations to produce more complex effects.

\begin{figure}[t]
    \centering
    \includegraphics[width=\linewidth]{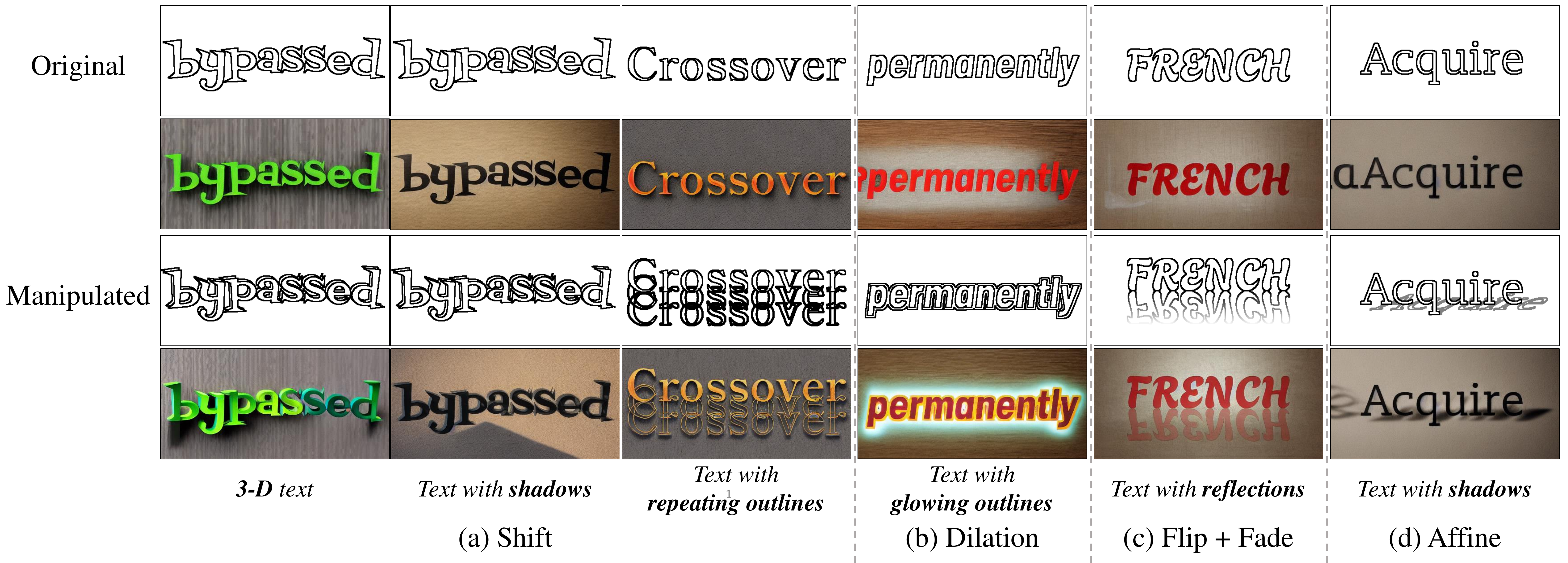}\\[-3mm]
    \caption{Examples of text images generated with effects by edge manipulation. Edges shown are dilated for better visualization. Please zoom in to confirm the detailed edges.}
    \label{fig:generated-with-effects}
\end{figure}

\subsubsection{Examples of Generated Texts with Edge Manipulations}
We present the results of generated texts with and without edge manipulations in Fig.~\ref{fig:generated-with-effects}. We observe that results generated with edge manipulations effectively incorporate the specified effects into the texts. Conversely, results generated without edge manipulations show minimal to no effects on the texts, even though the prompts given are the same. 
By specifying the amount of shift, we demonstrate the ability to control the width and direction of the 3-D and shadow effects (a) (left, middle). 
Furthermore, we extend the application of shift by obtaining the edges of the shifted edges, thereby achieving a repeating outline effect in (a) (right). 
In (b), the edge of the dilated text serves as a guide for ControlNet to produce outlines that accurately follow the original text.
By fading the flipped edges, we achieved a realistic reflection effect where the reflections are blurred, as depicted in (c).
Whereas (d) illustrates an example of a blurred shadow effect when the strength multiplier of the rotated edge is set to 0.5.

\section{Text Generation on Predefined Background}\label{sec:text-editing}

\begin{figure}[t]
    \centering
    \includegraphics[width=0.9\linewidth]{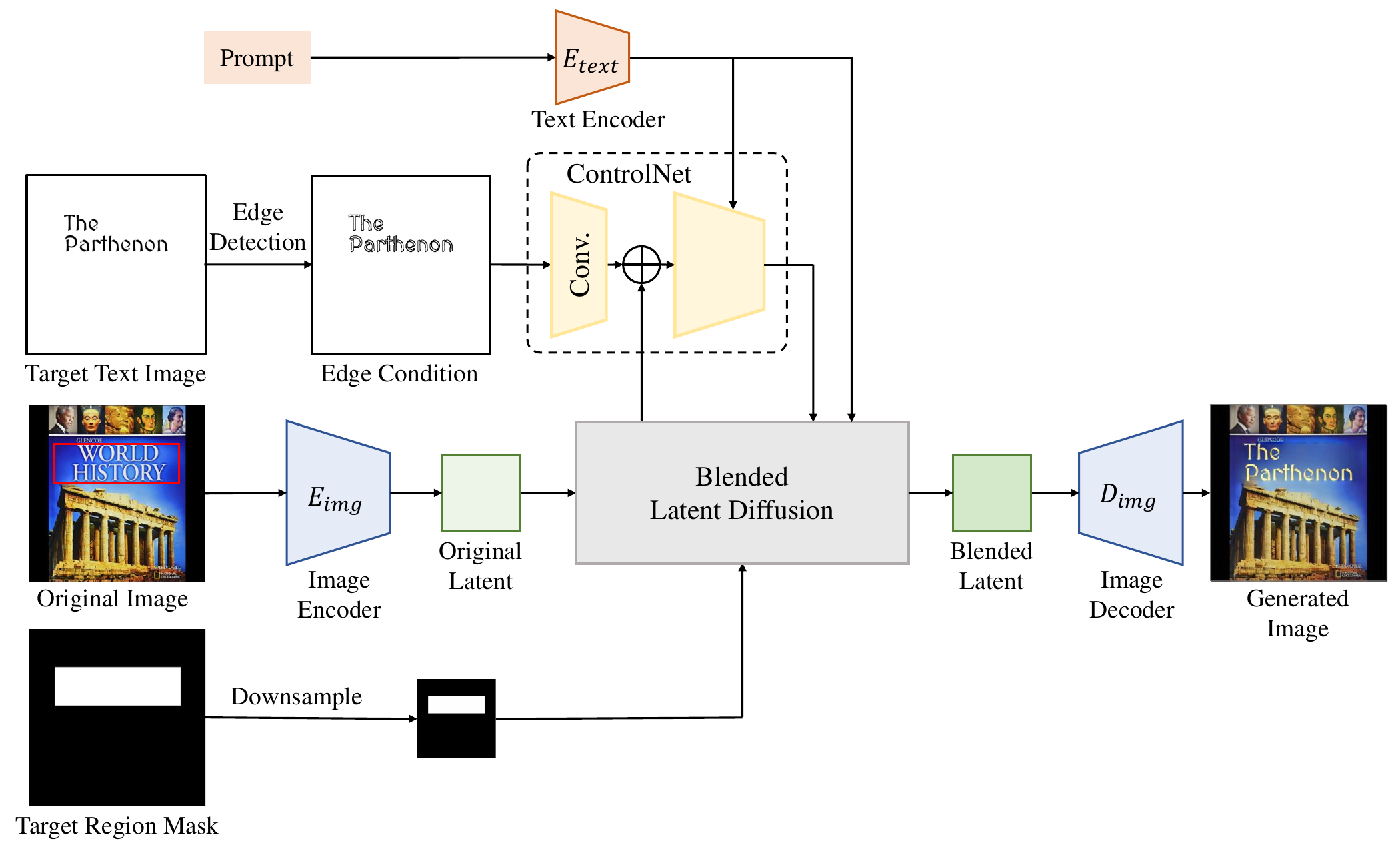}\\[-3mm]
    \caption{Overview of the proposed typographic text generation framework. The edge condition can be extracted from existing texts or rendered texts specified by the user.}
    \label{fig:overview_editing}
\end{figure}

\subsection{Combining Blended Latent Diffusion with ControlNet for Typographic Text Generation} \label{sec:editing_overview}

Blended Latent Diffusion~\cite{avrahami2023blended}  is a zero-shot approach to modify a part of an existing image while preserving the rest of the image using a pre-trained LDM. 
By providing a binary mask to indicate the region to be modified and the region to be preserved, the approach achieves seamless partial image generation through the repeated blending of the latents of the two regions. 
More specifically, at each time step of the denoising process, blending is carried out by replacing the region to be preserved in the noisy latent with a version of the latent of the original image noised to the current noise level. 
The LDM then denoises the blended noisy latent while preserving the overall coherence, leading to the modified region harmonizing with the original image.

We leverage the capability of Blended Latent Diffusion to naturally modify a part of an image and combine it with ControlNet's strengths in generating text to achieve text generation within typographic designs. 
The overview of our proposed system is illustrated in Fig.~\ref{fig:overview_editing}. 
Providing an image and a binary mask specifying the target region to be modified, we first obtain an edge condition from a target text that is within the masked area. 
Next, the original image is encoded into latent space, while the target region mask is downsampled to the dimensions of the latent to be used for blending.
Finally, blending is performed on the original latent while accepting the edge condition from ControlNet together with a text prompt, resulting in a generated text that naturally blends with the original image.

\subsection{Data Preparation for Evaluation} \label{sec:editing_dataset}
We compared the performance of the proposed typography text generation system with TextDiffuser quantitatively and qualitatively by generating texts on the publicly available typographic image dataset of book covers~\cite{iwana2017judging}. From the dataset containing 207,000 book covers, we randomly selected 1,000 and performed OCR to identify text regions.
Subsequently, for each book cover, we performed text generation 4 separate times, resulting in a total of 4,000 modified book covers.
Each time, a new text is generated onto a random text region that is at least 40 pixels in height and 100 pixels in width on the book cover.
Using the bounding boxes of the selected text regions as target region masks, we rendered a random text in a random font within each masked region to be used as the target text.
The target texts were then utilized to extract edge conditions for our method and generate text segmentation maps for TextDiffuser.

\begin{figure}[t]
    \centering
    \includegraphics[width=0.9\linewidth]{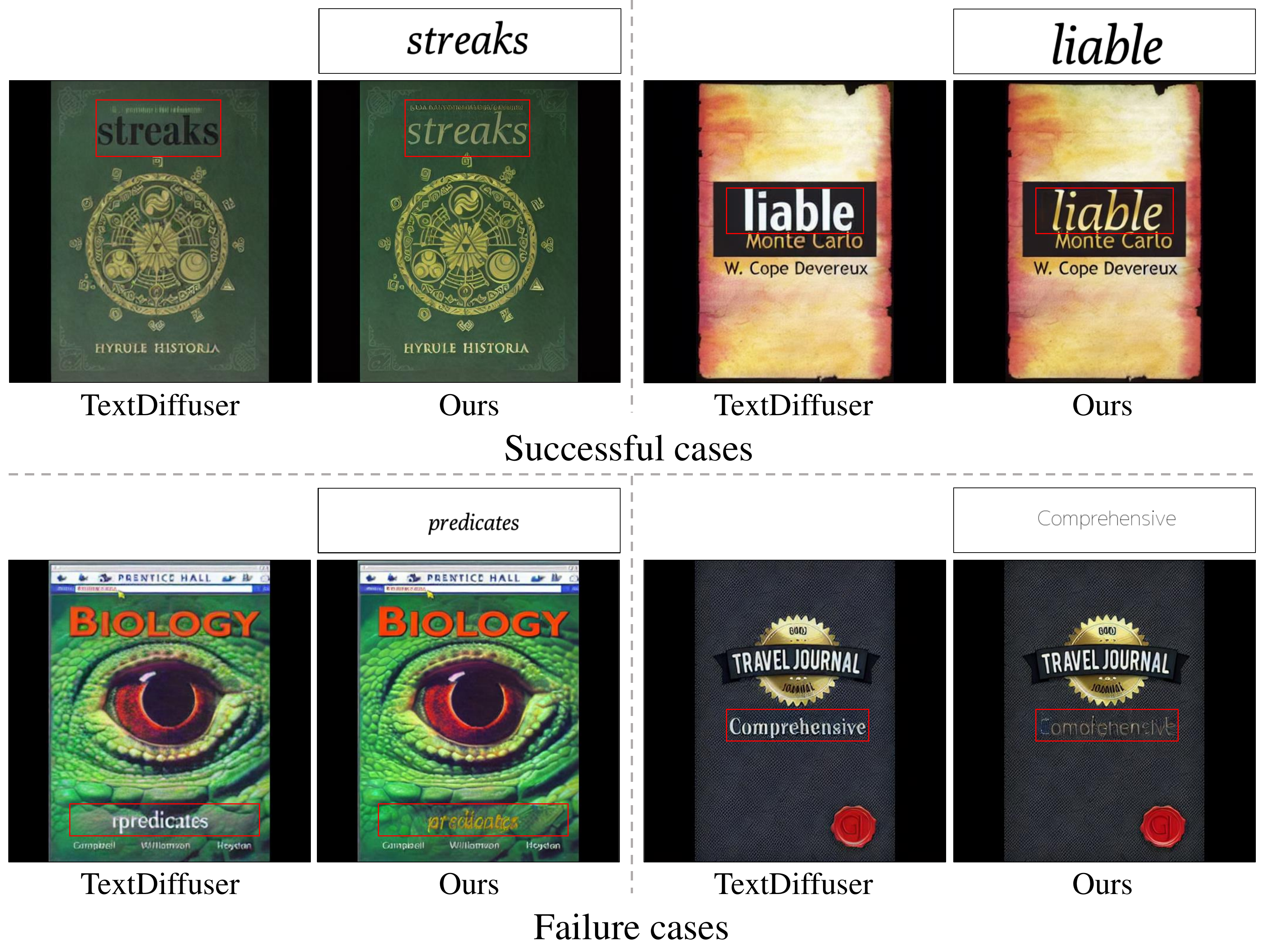}\\[-3mm]
    \caption{Comparison of typographic text generation results. Prompts used are ``\textit{Texts that read {\tt [word]} on a book cover}'' for all examples. The top row shows successful cases, while the bottom row shows failure cases.}
    \label{fig:comparison}
\end{figure}
\subsection{Qualitative Evaluation of Typographic Text Generation}  \label{sec:editing_qualitative}

Fig.~\ref{fig:comparison} shows some results of text generation on book cover background images using both TextDiffuser and the proposed method. 
We demonstrate that the proposed method can generate texts coherently with the background while reflecting the specified font styles, just like the experimental results for single-word generation. As shown in Table~\ref{tab:func-comparison}, TextDiffuser is capable of generating certain texts on a given background image. However, TextDiffuser does not consider the style of the texts, and therefore, it generates the texts in a style inferred from the background image. 
Consequently, even if a user intends to put the word ``streaks'' in an Italic style, there is no way to tell the style to TextDiffuser and the resulting text is printed in a non-Italic style. As noted before, in practical typographic workflows, it is common for book-cover designers to have specific ideas on text styles (by considering the content and genre of the book). The inability to accommodate these ideas is a fatal drawback in such workflows.
\par

The bottom row of Fig.~\ref{fig:comparison} displays failure cases of the proposed method. Compared to the previous experiment on single-word generation, the specified text height often has fewer pixels (40 pixels height in minimum). Then, the text edges from an ultra-light font style become difficult to work as a reasonable condition for ControlNet. Consequently, the generated text images differ from what the user specified. We found that most failures occur when texts are small.

\subsection{Quantitative Evaluation of Typographic Text Generation}   \label{sec:editing_quantitative}
\subsubsection{Evaluation Metrics} 
We quantitatively compared the proposed method with TextDiffuser using Fréchet Inception Distance (FID)~\cite{heusel2017gans} and OCR metrics outlined in Section~\ref{sec: ocr-metrics}. 
We employed FID to assess the quality of generated images compared to the distribution of original book covers.
FID was measured on both whole images and cropped versions, where the cropped versions contained the generated text regions along with a margin around it with a size ratio of 1:1. 
For OCR, evaluation is conducted solely on the generated regions.

\subsubsection{Evaluation Results}
Table~\ref{table:editing-quantitative} shows the result of the comparative experiment. Overall, these quantitative evaluation results do not show a large difference between ours and TextDiffuser, even though TextDiffuser was trained to generate texts on user-specified backgrounds. Note that for both ours and TextDiffuser, the OCR performance indexes in Table~\ref{table:editing-quantitative} are degraded from those in Table~\ref{table:generation-quantitative}. The reason for this degradation is 
that both methods struggle to generate high-quality character images when the text height is small.

\begin{table}[t]
    \centering
    \caption{Quantitative evaluation of typographic text generation.}
    \label{table:editing-quantitative}
    \begin{tabular}{c|cc|cccc} \hline
                & \multicolumn{2}{c}{Quality} & \multicolumn{4}{|c}{Legibility (OCR)} \\
         Method & FID(whole)$\downarrow$ & FID(cropped)$\downarrow$ & Precision$\uparrow$   & Recall$\uparrow$  & F1-score$\uparrow$ & NED$\uparrow$ \\ \hline
         TextDiffuser   & 15.09         & 13.99                     & 0.6640                & 0.6830            & 0.6701            & 0.9091 \\
         Ours           & 14.04         & 19.38                     & 0.6201                & 0.6565            & 0.6317            & 0.8851  \\ \hline
    \end{tabular}
\end{table}

\section{Further Application: Add Post-effects to Existing Texts} \label{sec:editing_w/effects}

Our proposed method shows its strengths in generating and adding customizable effects not only to user-specified fonts but also to existing typographic texts. 
To demonstrate this, we selected book covers from the same dataset mentioned in Section~\ref{sec:editing_dataset} and added post-effects to the texts on the book covers.
From an original book cover, we extracted edges of the target text and applied manipulations as described in Section~\ref{sec:manipulation_generation} to obtain the manipulated edges. 
Then, by using the manipulated edges as conditions together with a matching prompt, we add post-effects to existing texts.

\begin{figure}[t]
    \centering
    \includegraphics[width=0.79\linewidth]{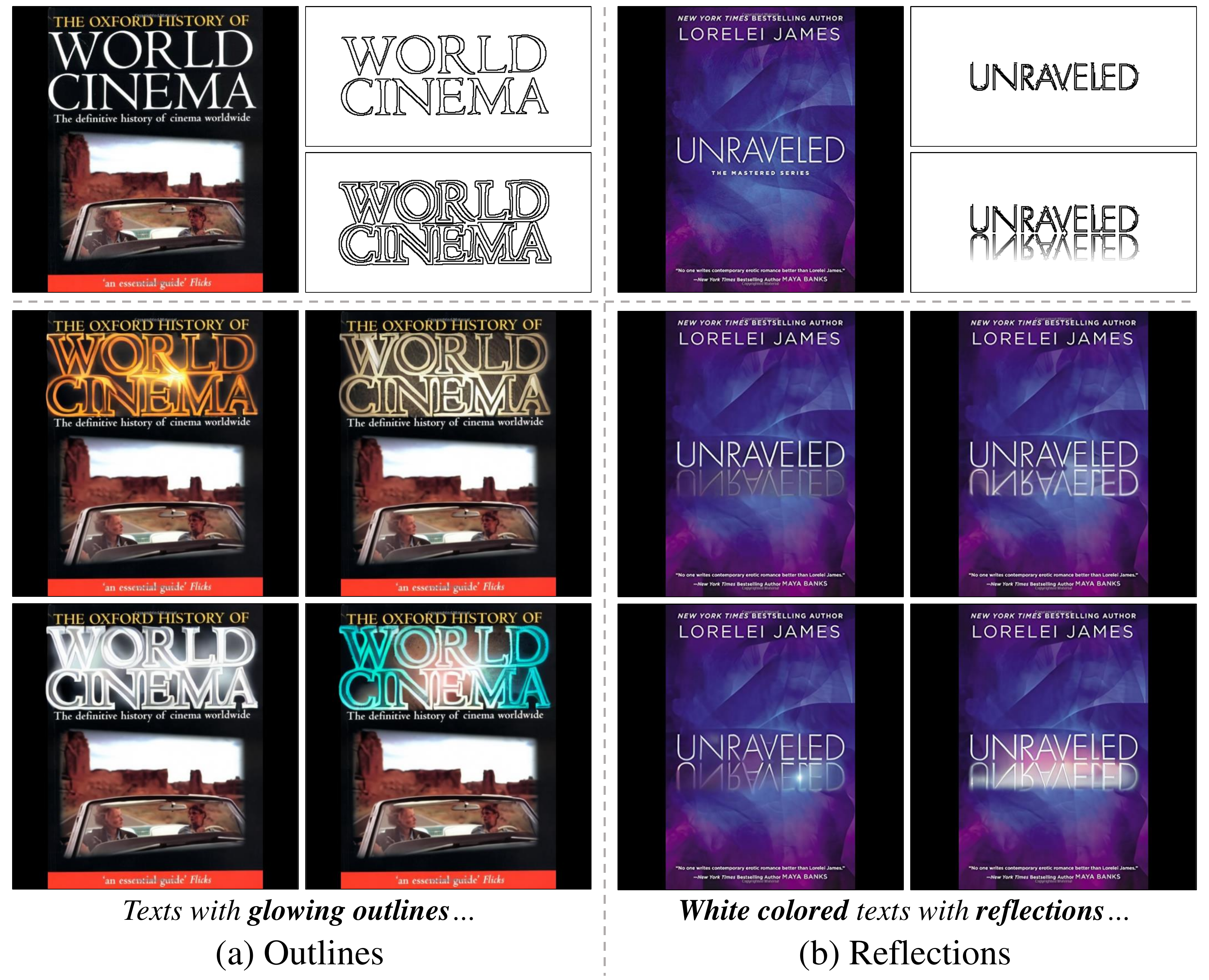}\\[-3mm]
    \caption{Examples of post-effects on typographic texts. Effects are added to texts in the original images displayed at the top left of each subfigure. On the right of each original image are the edges of the original text (top) and the manipulated edges (bottom). We show that our proposed method can generate multiple results with the same inputs.}
    \label{fig:text-editing}
\end{figure}

Fig.~\ref{fig:text-editing} displays examples of adding post-effects to existing texts on typographic images. On the top left of each subfigure is the original book cover. Edges are on the right of the original book covers, where the edges of the original texts are displayed above the manipulated edges.
We showcase that our method can introduce various effects to existing texts.
In (a), we used ``dilation'' to produce a glowing outline effect, while in (b), ``flip and fade'' were used to produce a reflection effect.
The above results demonstrate that as long as the edges can be obtained, our system is capable of adding post-effects to rasterized typographic texts without the need for a vector font file.
Furthermore, the ability to generate multiple results with the same inputs provides designers with choices to suit their needs.

\section{Conclusion}
We proposed a typographic text generation system that satisfies the practical requirements of the typographic process through a novel combination of two off-the-shelf methods for diffusion models, ControlNet and Blended Latent Diffusion. 
We demonstrated that ControlNet is ideal for generating high-quality typographic texts, as its ability to follow prompts and faithfully adhere to edge conditions enables users to explicitly specify the font styles, colors, and effects of the texts. 
Additionally, we experimentally proved that manipulating text edges allows for a more intuitive and controllable way to produce complex text effects.
Finally, we successfully showed that our proposed system is capable of adding and generating text on a predefined background while preserving its overall coherence.

As for the limitations and failure cases, our system struggles to generate extremely small and fine texts, as it is constrained by the upper bound of the VAE's reconstruction abilities.
Moreover, the system occasionally generates unwanted texts in unintended positions.
Future work includes addressing the limitation of generating small-scale texts and constraining the generation of texts to intended positions only.


\par
\noindent{\bf Acknowledgment}:\ This work was supported by JSPS KAKENHI Grant Number JP22H00540.
%
\bibliographystyle{splncs04}
\bibliography{sample-base}

\end{document}